\documentclass[letterpaper, 10 pt, journal, twoside]{IEEEtran}
\usepackage{xcolor}
\usepackage{amsmath}
\usepackage{amsfonts}
\usepackage[noend]{algpseudocode}
\usepackage[ruled,linesnumbered]{algorithm2e}
\SetArgSty{textup}
\usepackage{amssymb}
\usepackage{mathtools}
\usepackage{mathrsfs}
\usepackage{multirow}
\usepackage{indentfirst}
\usepackage{booktabs}
\usepackage{tabularx}
\usepackage{threeparttable}
\usepackage[font=small]{caption}
\usepackage[font=small]{subcaption}
\usepackage{array}
\usepackage{textcomp}
\usepackage{stfloats}
\usepackage{url}
\usepackage{verbatim}
\usepackage{graphicx}
\usepackage{amsthm}
\usepackage{cite}
\usepackage{hyperref}
\usepackage{makecell}
\usepackage{diagbox}
\usepackage{tikz}

\hypersetup{
    colorlinks=true,
    linkcolor=red,
    citecolor=red,
    urlcolor=blue
}

\def\BibTeX{{\rm B\kern-.05em{\sc i\kern-.025em b}\kern-.08em
    T\kern-.1667em\lower.7ex\hbox{E}\kern-.125emX}}
\usepackage{balance}

\usepackage{amsthm}
\theoremstyle{definition}

\newtheorem{lemma}{\bf Lemma}

\usepackage{xcolor}

\begin{document}

\title{DMVC-Tracker: Distributed Multi-Agent Trajectory Planning for Target Tracking Using Dynamic Buffered Voronoi and Inter-Visibility Cells}

\author{Yunwoo Lee$^{1}$, Jungwon Park$^{2}$, and H. Jin Kim$^{3}$
\thanks{Manuscript received: November 4, 2024; Revised: January 31, 2025; Accepted: March 3, 2025}\
\thanks{This paper was recommended for publication by Editor M. Ani Hsieh upon
evaluation of the Associate Editor and Reviewers’ comments.
This work was supported by the IITP-ITRC (IITP-2025-RS-2024-00437268).
}
\thanks{$^{1}$The author is with Artificial Intelligence Institute of Seoul National University (AIIS), Seoul, South Korea (e-mail: snunoo12@snu.ac.kr).}\
\thanks{$^{2}$The author is with the Department of Mechanical System Design Engineering, Seoul National University of Science and Technology, Seoul, South Korea (e-mail: jungwonpark@seoultech.ac.kr).}\
\thanks{$^{3}$The author is with the Department of Aerospace Engineering, Seoul National University, Seoul, South Korea (e-mail: hjinkim@snu.ac.kr).}
\thanks{Digital Object Identifier (DOI): see top of this page.}
}


\markboth{IEEE Robotics and Automation Letters. Preprint Version. Accepted March, 2025}{Lee \MakeLowercase{\textit{et al.}}: DMVC-Tracker: Distributed Multi-Agent Trajectory Planning for Target Tracking}

\maketitle
\begin{abstract}
This letter presents a distributed trajectory planning method for multi-agent aerial tracking. The proposed method uses a Dynamic Buffered Voronoi Cell (DBVC) and a Dynamic Inter-Visibility Cell (DIVC) to formulate the distributed trajectory generation. Specifically, the DBVC and the DIVC are time-variant spaces that prevent mutual collisions and occlusions among agents, while enabling them to maintain suitable distances from the moving target. We combine the DBVC and the DIVC with an efficient Bernstein polynomial motion primitive-based tracking trajectory generation method, which has been refined into a less conservative approach than in our previous work. The proposed algorithm can compute each agent's trajectory within several milliseconds on an Intel i7 desktop. We validate the tracking performance in challenging scenarios, including environments with dozens of obstacles.
\end{abstract}

\vspace{-3mm}

\IEEEpeerreviewmaketitle

\section{Introduction}
\label{sec:introduction}
\IEEEPARstart{A}{erial} target tracking has been widely applied in fields such as cinematography and surveillance. Although a single micro aerial vehicle (MAV) is usually employed in these applications, the utilization of multiple MAVs can bring benefits. For example, multiple views of actors captured by a team of MAVs provide movie directors with more footage \cite{doyouseewhatisee, alcantara,lighting}. Also, multi-agent tracking can be deployed in moving motion-capture systems \cite{mocap1,mocap2,mocap3}, and a large number of cameras increases the accuracy of pose estimation.

Despite great attention and research on motion generation for multi-robot systems, multi-agent target tracking remains a challenging task. The main challenge is finding constraints that prevent both inter-agent occlusion and inter-agent collision, while considering the moving target. Additionally, several other requirements should be considered: occlusion and collision against obstacles, actuator limits, and tracking distances. To respond to frequent changes in the motion of the target and moving obstacles, the trajectory planning should be updated quickly to reflect these considerations.
\begin{figure}[t!]
\centering
\includegraphics[width = 0.8\linewidth]{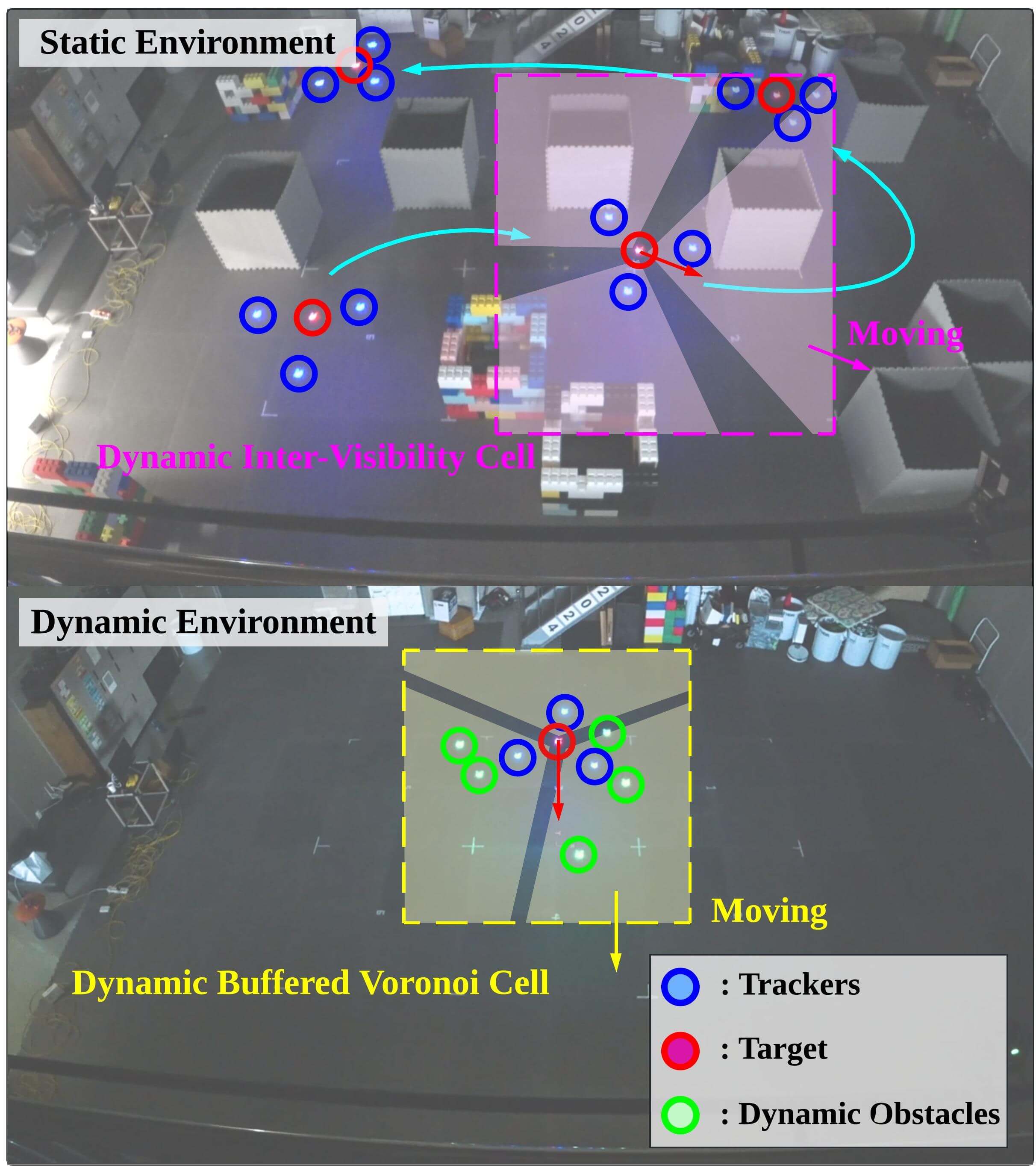}
\caption{Hardware demonstration of multi-agent target tracking.}
\label{fig:thumbnail}
\vspace{-6mm}
\end{figure}

In this letter, we present an online distributed trajectory planning algorithm that can generate a target-visible and safe trajectory. The key ideas of the proposed method are Dynamic Buffered Voronoi Cell (DBVC) and Dynamic Inter-Visibility Cell (DIVC). 
The DBVC is a time-varying, inter-collision-free region developed from  Buffered Voronoi Cell (BVC) \cite{buffered_voronoi_cell}.
This time-varying region helps maintain consistent distances between the target and the agents, whereas the BVC can cause an agent to become stuck in a static space, making it difficult for the agent to follow the target. We design the DIVC to prevent inter-agent occlusion. Specifically, the DIVC ensures that the \textit{Line-of-Sight} connecting the target and an agent does not collide with other agents.
Based on the DBVC and the DIVC, the proposed planner generates a target-visible and safe trajectory using Bernstein polynomial motion primitives. We employ a \textit{sample-check-select} strategy and leverage properties of Bernstein polynomials to speed up the calculation, enabling us effectively to tackle complex tracking problems. This paper reduces the conservativeness of the \textit{check} method from our previous work \cite{bpmp-tracker} to achieve a higher success rate.
The main contributions of this paper are as follows.
\begin{itemize}
    \item A distributed multi-agent trajectory planning algorithm for target tracking that generates a collision-free and occlusion-free trajectory in complex environments, such as dozens of obstacles.
    \item Dynamic Buffered Voronoi Cell (DBVC) and Dynamic Inter-Visibility Cell (DIVC) that impose inter-agent collision and inter-agent occlusion avoidance constraints, while adapting to the future movements of the target.
    \item An integration of the DBVC and the DIVC with an improved Bernstein polynomial motion primitive-based trajectory generator, which lessens the conservativeness of our previous method \cite{bpmp-tracker}.
\end{itemize}

\section{Related Works}
\label{sec:related_works}
\subsection{Target Tracking Using a Single Drone}
\label{subsec:rel_work_single}
Various studies have proposed methods that consider target visibility for single-drone path planning in obstacle environments. \cite{elastic_tracker,auto_filmer,boseong_icra} address target occlusion in the presence of static and unstructured obstacles. To be specific, \cite{elastic_tracker} and \cite{auto_filmer} design polytope-shaped target-visible regions and formulate a spatio-temporal optimization problem. \cite{boseong_icra} proposes a visibility metric using Euclidean Signed Distance Fields (ESDF) and formulates a graph optimization to select optimal viewpoints. 

There are works that address target tracking problems in dynamic obstacle environments \cite{nageli,multi-convex, qp_chaser}. \cite{nageli} designs a visibility cost to avoid target occlusion, inspired by the GPU ray casting model. \cite{multi-convex} utilizes partial convex structures of the target tracking optimization problem to handle non-convex visibility constraints, while \cite{qp_chaser} designs a time-varying half-space-shaped region to apply visibility constraints within a quadratic programming problem.

The above approaches can be utilized for multi-agent target tracking missions by treating neighboring drones as dynamic obstacles. However, such approaches may result in collisions between agents or occlusions of the target by neighboring agents, as the newly updated trajectories may differ from the predicted movements of other agents. In contrast, our planner enables cooperative target tracking, in a \textit{distributed} manner.

\subsection{Target Tracking Using Multiple Drones}
\label{subsec:rel_work_multiple}
Research on target tracking using multiple MAVs has been widely studied in recent years. \cite{mocap1,mocap2,lighting} focus on the formation of MAVs to follow the moving target. \cite{lighting} and \cite{mocap1} decompose the multi-MAV trajectory optimization to tackle non-convex constraints, which allows replanning in real-time. Both methods calculate viewpoints considering formations, and then successive optimization generates a safe \cite{lighting} and smooth \cite{mocap1} trajectory. \cite{mocap2} develops external control inputs to prevent inter-agent collision, enabling the conversion of the optimization problem into a convex one. However, the above approaches address only collision avoidance with respect to the target and neighboring agents, while omitting considerations of visibility issues caused by neighboring agents.

On the other hand, there are studies that consider mutual visibility. \cite{doyouseewhatisee} optimize MAVs' viewpoints in discretized state and time spaces in a centralized manner. The finer the discretized states, the more computation time and memory are required. Therefore, the authors set the number of time steps small to enable real-time execution of the algorithm. The methods in \cite{alcantara} and \cite{multi-view} try to avoid the appearance of the other MAVs in the MAV's camera image.
Both methods operate in a distributed manner to meet real-time criteria; however, \cite{alcantara} uses a priority-based approach, while \cite{multi-view} employs a sequential consensus approach. Therefore, there may be situations where all mutual visibility constraints are not simultaneously satisfied. 
In contrast, our method ensures inter-agent occlusion avoidance and does not require sequential planning updates, resulting in a fully distributed approach.

\subsection{Target Tracking in Crowded Environments}
\label{subsec:discussion_efficient}
Most trajectory planning research aims to ensure real-time performance. 
The works \cite{nageli}, \cite{bonatti}, and \cite{penin} formulate non-convex optimization methods to generate a target-visible trajectory; however, as the number of obstacles increases, computation time quickly increases.
To address this, this work employs Bernstein polynomial motion primitives to efficiently find a tracking trajectory by leveraging the properties of Bernstein polynomials. This work reduces the conservativeness of the approach in our previous work \cite{bpmp-tracker}, further improving the tracking success rate in challenging conditions, such as environments with numerous dynamic obstacles.
\begin{table}[t]
    \vspace{1mm}
    \caption{Nomenclature}
    \centering
    \label{tab:nomenclature}
        \begin{tabular}{|c|c|}
        \hline
        \textbf{Symbol} & \textbf{Definition}\\
        \hline
        $\textbf{x}_{c}^{i}(t)$  & Trajectory of the $i$-th agent. \\
        \hline
        $\textbf{x}_{q}(t)$  & Predicted trajectory of the target.\\
        \hline
        $\textbf{x}_{o}^{k}(t)$  & Predicted trajectory of the $k$-th dynamic obstacle.\\
        \hline
        $N_{c}, N_{o}$  & The number of agents and dynamic obstacles. \\
        \hline
        $\mathcal{I}_{c}$  & An agent index set. $\mathcal{I}_{c}=\{1,\ldots,N_{c}\}$\\
        \hline
        $\mathcal{I}_{o}$  & A dynamical obstacle index set. $\mathcal{I}_{o}=\{1,\ldots,N_{o}\}$\\
        \hline
        $r_{c}, r_{o}, r_{q}$  & Radius of agents, dynamic obstacles and a target. \\
        \hline
        $T$ & Planning horizon.\\
        \hline
        $\mathcal{X}$, $\mathcal{T}(t)$ &Environment, Occupied space by the target.\\
        \hline
        $\mathcal{C}_{i}(t), \mathcal{C}(t)$ & Occupied space by the $i$-th agent and all agents.\\
        \hline
        $\mathcal{C}^{i}(t)$ & Occupied space excluding $\mathcal{C}_{i}(t)$ from $\mathcal{C}(t)$\\
        \hline
        $\mathcal{F}_{s},\mathcal{F}_{d}(t)$ & Static-obstacle-free space, obstacle-free space\\
        \hline
        $\mathcal{F}_{q}(t)$ & The space in $\mathcal{F}_{d}(t)$ excluding the target's area.\\
        \hline
        $\mathcal{F}_{c}^{i}(t)$ & The space in $\mathcal{X}$ excluding $\mathcal{C}^{i}(t)$.\\
        \hline
        $\mathcal{H}_{s}^{ij}(t)$ & DBVC for the $i$-th agent against the $j$-th agent.\\
        \hline
        $\bigcap_{k=1,2}\mathcal{H}_{\mu k}^{ij}(t)$, & DIVC for the $i$-th agent against the $j$-th agent.\\
        $\mu= o$ or $a$ & $o$: obtuse case, $a$: acute case. \\
        \hline
        $\mathcal{B}(\textbf{x},r)$ & A ball with center at $\textbf{x}$ and radius $r$.\\
        \hline
        $\mathcal{L}(\textbf{x}_{a}, \textbf{x}_{b})$ & A segment connecting points at $\textbf{x}_{a}$ and $\textbf{x}_{b}$.\\
        \hline
        $\cup, \cap, \setminus $ & Set operator: union, intersect, except\\
        \hline
        $\oplus$& Minkowski-sum operator\\
        \hline
        $\|\textbf{x}\|, \textbf{x}^{\top}$ & Euclidean norm and transpose of a vector $\textbf{x}$.\\
        \hline
        $[A;B]$ & Row-wise concatenation of matrices $A$ and $B$\\
        \hline
        $\overline{AB}$ & A segment connecting points $A$ and $B$.\\
        \hline
        \end{tabular}
        \vspace{-4mm}
\end{table}
\section{Problem Formulation}
\label{sec:problem_formulation}
In this section, we formulate a trajectory planning problem for multiple tracking agents. We suppose that $N_{c}$ homogenous agents are deployed to follow a target in a 2-dimensional space $\mathcal{X}\subset \mathbb{R}^{2}$ with static and $N_{o}$ dynamic obstacles. We aim to generate trajectories so that the agents 1) avoid collisions, 2) maintain the visibility of the target and 3) do not exceed dynamical limits. 

For consistent image acquisition of the target, we fix the flying altitude, which means we address the 2D problem. Unlike methods that allow drones to move obstacle-free high altitudes \cite{obstacle_free3d1,obstacle_free3d2}, since the drone must find feasible movements only on the $x$-$y$ plane, our problem is more challenging. Throughout this paper, the notations in Table \ref{tab:nomenclature} are used.

\subsection{Environments}
\label{subsec:environments}
The environment $\mathcal{X}$ consists of target space $\mathcal{T}(t)$, obstacle space $\mathcal{O}(t)$, and drone space $\mathcal{C}(t)$. $\mathcal{O}(t)$ is divided into spaces occupied by static and dynamic obstacles, $\mathcal{O}_{s}$ and $\mathcal{O}_{d}(t)$, respectively. $\mathcal{C}(t)$ is an occupied space by all agents and is partitioned into the $i$-th agent space $\mathcal{C}_{i}(t)$, $i \in \mathcal{I}_{c}$.
We define the following free spaces for target-tracking missions.
\begin{equation}
    \label{eq:free_spaces}
\begin{aligned}
    &\mathcal{F}_{s} = \mathcal{X}\setminus \mathcal{O}_{s}, \ 
     \mathcal{F}_{d}(t) = \mathcal{F}_{s}\setminus \mathcal{O}_{d}(t),\
     \mathcal{F}_{q}(t) = \mathcal{F}_{d}(t)\setminus \mathcal{T}(t),\\
    &\mathcal{F}_{c}^{i}(t) = \mathcal{X}\setminus \mathcal{C}^{i}(t), \ \text{where}\ \mathcal{C}^{i}(t) = \mathcal{C}(t) \setminus \mathcal{C}_{i}(t).
\end{aligned}
\end{equation}

\subsection{Trajectory Representation}
\label{subsec:trajectory_representation}
Due to the virtue of differential flatness of the dynamics of various MAV platforms, such as quadrotors, we represent trajectories of all agents as polynomial functions of time $t$. Employing the Bernstein basis, we represent the trajectory of  the $i$-th agent, $\textbf{x}^{i}_{c}(t)\in \mathbb{R}^{2}, i\in \mathcal{I}_{c}$ as follows:
\begin{equation}
    \textbf{x}_{c}^{i}(t) = \textbf{C}^{i\top}\textbf{b}_{n_{c}}(t),\ t \in [0,T]
    \label{eq:berstein_representation}
\end{equation}
where $T$ is the planning horizon, $n_{c}$ is the degree of the polynomials, $\textbf{C}^{i}\in\mathbb{R}^{(n_c+1)\times2}$ is a coefficient matrix for the agent $i$, and $\textbf{b}_{n_{c}}(t)\in \mathbb{R}^{(n_{c}+1)\times1}$ is a vector that consists of $n_{c}$-th order Bernstein bases for time interval $[0,T]$. We represent the trajectory of the target, $\textbf{x}_{q}(t)$, the $k$-th dynamic obstacles, $\textbf{x}_{o}^{k}(t)$, $k\in \mathcal{I}_{o}$ in the same manner.

\subsection{Assumptions}
\label{subsec:assumptions}
In this study, we make the following assumptions.
\begin{itemize}
    \item \textit{Agents}: All agents share their current positions, and they start trajectory planning at the same time.
    \item \textit{Obstacles}: The information about the static obstacle space $\mathcal{O}_{s}$ is given as a point cloud, and the current positions of dynamic obstacles can be acquired.
    \item \textit{Target}: The current position of the target can be observed.
    \item \textit{Shape}: The moving objects, such as the target, dynamic obstacles, and tracking agents are modeled as balls with radius $r_{q}$, $r_{o}$ and $r_{c}$, respectively.
\end{itemize}
In this paper, we use extended Kalman filters to estimate the velocity of each moving object, and the future trajectories of the obstacles and the target are predicted using a constant velocity model and the method described in \cite{bpmp-tracker}, respectively.
\subsection{Mission Description}
\label{subsec:mission_description}
In the multi-agent tracking trajectory planning, we focus on the following objectives.
\subsubsection{Collision Avoidance}
\label{subsubsec:collision_avoidance}
For safety, all agents should not collide with obstacles, the target, and the other agents.
\begin{equation}
    \label{eq:collision_avoidance}
    \mathcal{B}(\textbf{x}_{c}^{i}(t),r_{c})\subset \mathcal{F}_{q}(t)\cap\mathcal{F}_{c}^{i}(t),\ \forall i\in \mathcal{I}_{c},\ \forall{t}\in [0,T]
\end{equation}

\subsubsection{Occlusion Avoidance}
\label{subsubsec:occlusion_avoidance}
To avoid target occlusion caused by obstacles or neighboring agents, the \textit{Lines-of-Sight} between the agents and the target should not intersect with any obstacles and other agents.
\begin{equation}
    \label{eq:occlusion_avoidance}
    \mathcal{L}(\textbf{x}_{c}^{i}(t),\textbf{x}_{q}(t))\subset \mathcal{F}_{d}(t)\cap\mathcal{F}_{c}^{i}(t),\ \forall i\in \mathcal{I}_{c},\ \forall{t}\in [0,T]
\end{equation}

\subsubsection{Target Distance}
\label{subsubsec:target_distance}
In order to avoid being too close or too far from the target, we formulate the distance constraints.
\begin{equation}
    \label{eq:distance_target}
    \|\textbf{x}_{c}^{i}(t)-\textbf{x}_{q}(t)\|\in [d_{\min}, d_{\max}],\ \forall i \in \mathcal{I}_{c},\
    \forall t \in [0,T]
\end{equation}
\subsubsection{Dynamical Limits}
\label{subsubsec:dynamical_limits}
Due to the actuator limits of the drone, the trajectory should not exceed the maximum velocity, $v_{\max}$, and acceleration, $a_{\max}$. Also, the yaw rate should not exceed $\dot{\psi}_{\max}$ to prevent motion blurs in the camera.
\begin{equation}
\label{eq:dynamical_limits}
    \begin{aligned}
        \|\dot{\textbf{x}}_{c}^{i}(t)\|\leq v_{\max},\ \|\ddot{\textbf{x}}_{c}^{i}(t)\|\leq a_{\max},\ &|\dot{\psi}_{c}^{i}(t)|\leq \dot{\psi}_{\max},\\
        &\forall i\in \mathcal{I}_{c},\ \forall t \in [0,T]
    \end{aligned}
\end{equation}
\section{Dynamic Cells for Multi-Agent Tracking}
\label{sec:moving_cells}
In this section, we design time-varying regions free from inter-agent collisions and inter-agent occlusions, which are suitable for dynamic target tracking.
\subsection{Dynamic Buffered Voronoi Cell}
\label{subsec:moving_buffered_voronoi_cells}
We consider the inter-collision-free region that moves along the target's predicted trajectory.
We define a Dynamic Buffered Voronoi Cell (DBVC) to avoid collisions between the $i$-th and $j$-th agents as follows:
\begin{subequations}
\label{eq:moving_buffered_voronoi_cells}
\begin{align}
    \nonumber \mathcal{H}_{s}^{ij}(t) = \Big\{ & \textbf{x}(t)\in \mathbb{R}^{2}| (\textbf{x}_{c0}^{j}-\textbf{x}_{c0}^{i})^{\top}\Big(\textbf{x}(t)-\textbf{x}_{q}(t)+\textbf{x}_{q0}-\\  & \frac{\textbf{x}_{c0}^{i}+\textbf{x}_{c0}^{j}}{2}\Big) + r_{c}\|\textbf{x}_{c0}^{j}-\textbf{x}_{c0}^{i}\| \leq 0 \Big\},\\
    \nonumber \mathcal{H}_{s}^{ji}(t) = \Big\{ & \textbf{x}(t)\in \mathbb{R}^{2}| (\textbf{x}_{c0}^{i}-\textbf{x}_{c0}^{j})^{\top}\Big(\textbf{x}(t)-\textbf{x}_{q}(t)+\textbf{x}_{q0}-\\  & \frac{\textbf{x}_{c0}^{j}+\textbf{x}_{c0}^{i}}{2} \Big) + r_{c}\|\textbf{x}_{c0}^{i}-\textbf{x}_{c0}^{j}\| \leq 0 \Big\}
\end{align}    
\end{subequations}
\begin{lemma}
\label{th:mbvc}
If $\textbf{x}_{c}^{i}(t)\in \mathcal{H}_{s}^{ij}(t)$ and $\textbf{x}_{c}^{j}(t)\in \mathcal{H}_{s}^{ji}(t)$, $i\neq j\in \mathcal{I}_{c}$, the $i$-th and $j$-th agents do not collide with each other. 
\end{lemma}
\begin{proof} When $\textbf{x}_{c}^{i}(t)\in \mathcal{H}_{s}^{ij}(t)$ and $\textbf{x}_{c}^{j}(t)\in \mathcal{H}_{s}^{ji}(t)$, the summation of constraints in (\ref{eq:moving_buffered_voronoi_cells}) yields:
\begin{equation}
    \label{eq:sum_mbvc}
    \begin{aligned}
    &(\textbf{x}_{c0}^{j}-\textbf{x}_{c0}^{i})^{\top}(\textbf{x}_{c}^{i}(t)-\textbf{x}_{c}^{j}(t))+2r_{c}\|\textbf{x}_{c0}^{j}-\textbf{x}_{c0}^{i}\|\leq0\\
    &\Leftrightarrow (\textbf{x}_{c0}^{i}-\textbf{x}_{c0}^{j})^{\top}(\textbf{x}_{c}^{i}(t)-\textbf{x}_{c}^{j}(t)) \geq 2r_{c}\|\textbf{x}_{c0}^{i}-\textbf{x}_{c0}^{j}\|
    \end{aligned}
\end{equation}
By using Cauchy-Schwartz Inequality and (\ref{eq:sum_mbvc}), we have
\begin{equation}
    \begin{aligned}
    \|\textbf{x}_{c}^{i}(t)-\textbf{x}_{c}^{j}(t)\| &\geq \frac{\|(\textbf{x}_{c0}^{i}-\textbf{x}_{c0}^{j})^{\top}(\textbf{x}_{c}^{i}(t)-\textbf{x}_{c}^{j}(t))\|}{\|\textbf{x}_{c0}^{i}-\textbf{x}_{c0}^{j}\|}\\
    &\geq \frac{2r_{c}\|\textbf{x}_{c0}^{i}-\textbf{x}_{c0}^{j}\|}{\|\textbf{x}_{c0}^{i}-\textbf{x}_{c0}^{j}\|}=2r_{c}.
    \end{aligned}
\end{equation}
Hence, we can conclude that $\mathcal{H}_{s}^{ij}(t)$ and $\mathcal{H}_{s}^{ji}(t)$ do not make inter-agent collisions. 
\end{proof}
\subsection{Dynamic Inter-Visibility Cell}
\label{subsec:moving_visibility_cells}
Similar to the Dynamic Buffered Voronoi Cell (DBVC), we define a Dynamic Inter-Visibility Cell (DIVC) that moves in accordance with the target's future trajectory and prevents inter-agent occlusion between the $i$-th and $j$-th agents. 
We design the DIVC by dividing the cases where an angle formed by the two \textit{Lines-of-Sight}, $\mathcal{L}(\textbf{x}_{c0}^{i},\textbf{x}_{q0})$ and $\mathcal{L}(\textbf{x}_{c0}^{j},\textbf{x}_{q0})$, is either obtuse or acute.
\subsubsection{Obtuse case}
\label{subsubsec:mvc_obtuse}
\begin{figure}[t!]
    \centering
    \begin{subfigure}[t]{0.242\textwidth}
    \centering \includegraphics[width=1.0\linewidth]{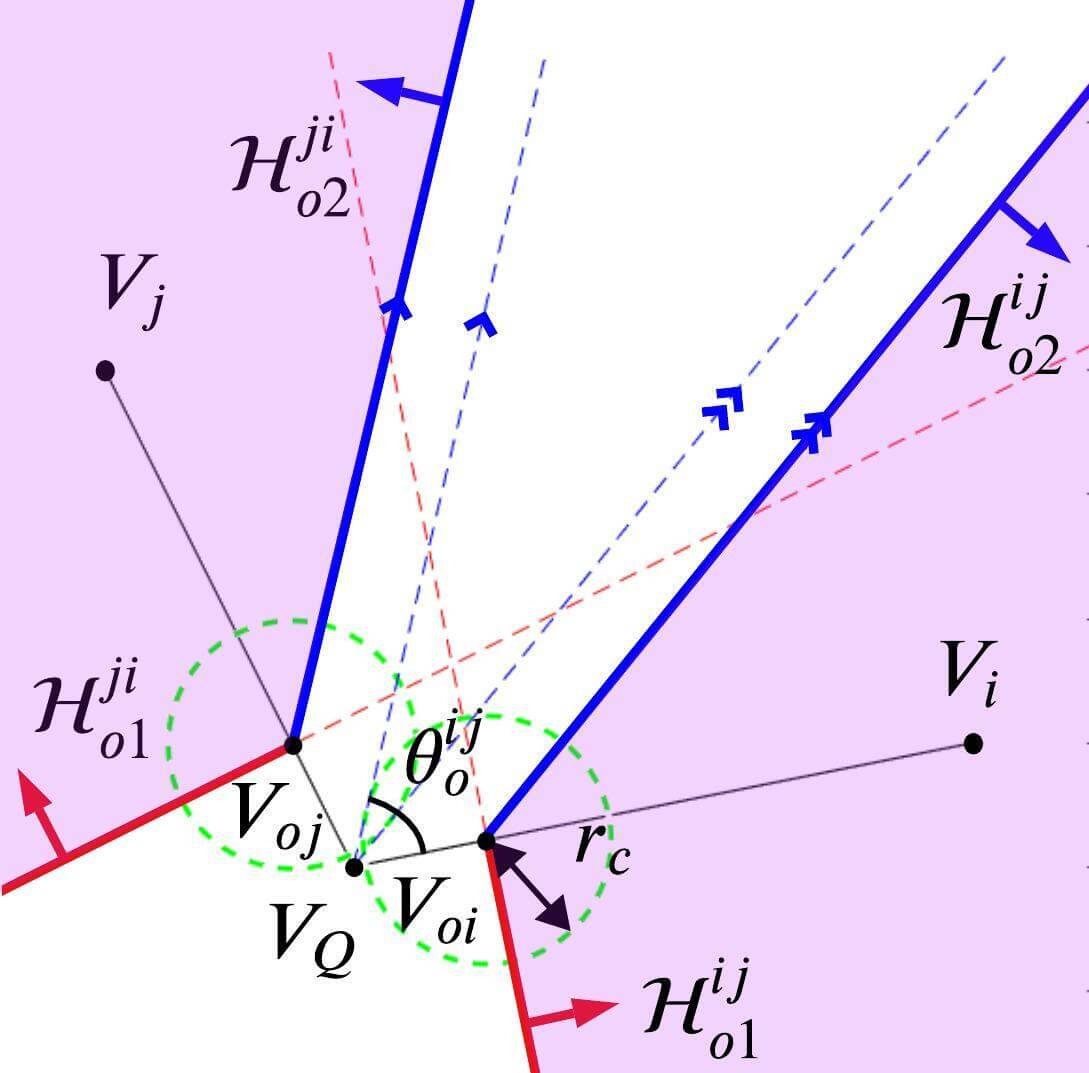}
    \caption{Obtuse case}
    \label{fig:constraint_mvc_obtuse}
    \end{subfigure}
    \begin{subfigure}[t]{0.240\textwidth}
    \centering \includegraphics[width=0.96\linewidth]{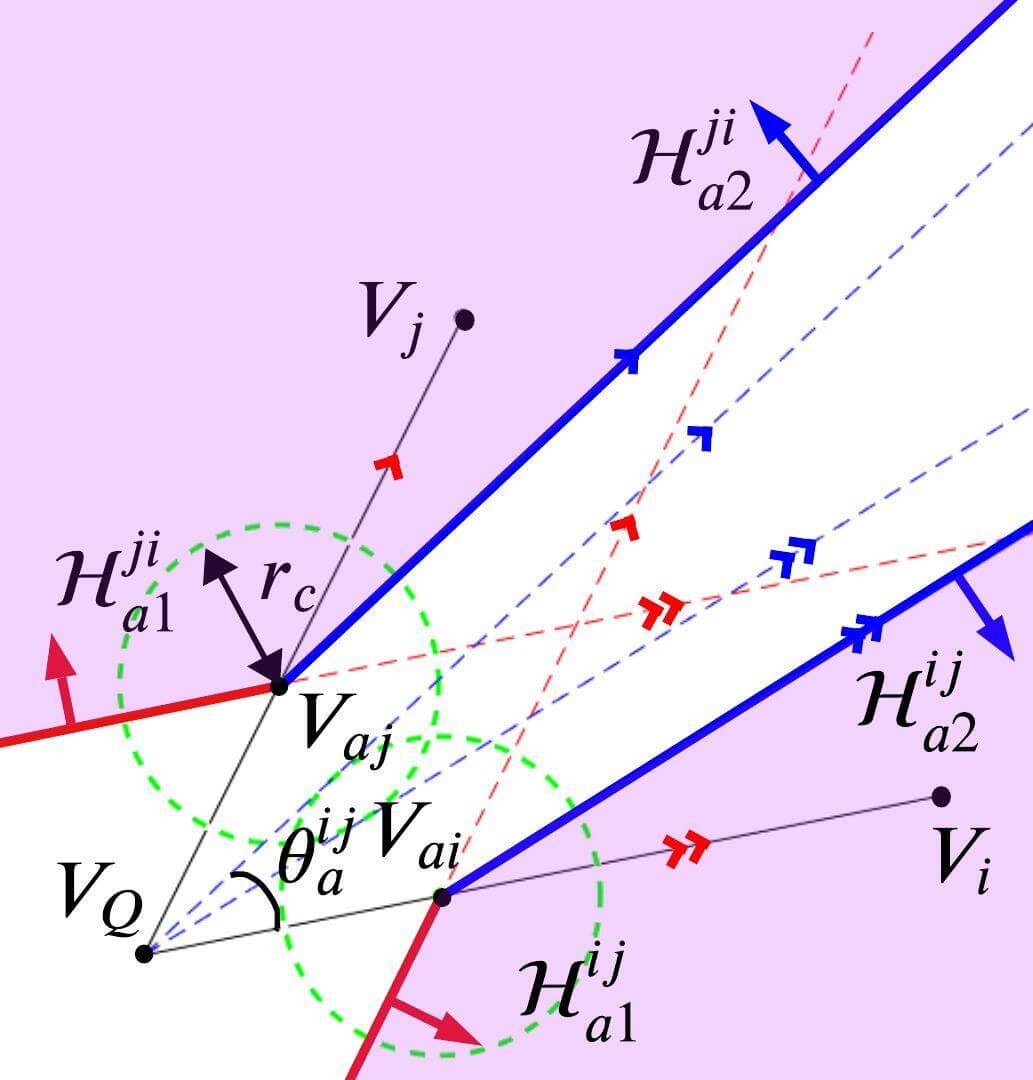}
    \caption{Acute case}
    \label{fig:constraint_mvc_acute}
    \end{subfigure}
    \caption{DIVC fomulation}
    \label{fig:constraint_mvc}
    \vspace{-4mm}
\end{figure}

Let $\textit{V}_{i}$, $\textit{V}_{j}$, and $\textit{V}_{Q}$ be points at $\textbf{x}_{c0}^{i}$, $\textbf{x}_{c0}^{j}$, and $\textbf{x}_{q0}$, respectively. 
Then, let points $V_{oi}$ and $V_{oj}$ be the points on line segments $\overline{V_{i}V_{Q}}$ and $\overline{V_{j}V_{Q}}$ that are each at a distance $\alpha_{o}^{ij}r_{c}$ ($\alpha_{o}^{ij}\geq1$) from $V_{Q}$. We draw lines (red in Fig. \ref{fig:constraint_mvc_obtuse}) that are perpendicular to $\overline{V_{i}V_{Q}}$ and $\overline{V_{j}V_{Q}}$, crossing $V_{oi}$ and $V_{oj}$, and we call half-spaces made by the lines $\mathcal{H}_{o1}^{ij}$ and $\mathcal{H}_{o1}^{ji}$, respectively. 
In addition, we draw the rays (blue in Fig. \ref{fig:constraint_mvc_obtuse}) starting from $V_{oi}$ and $V_{oj}$, which are parallel to the lines that pass through $V_{Q}$ and are tangential to the balls, $\mathcal{B}(\textit{V}_{oj},r_{c})$ and $\mathcal{B}(\textit{V}_{oi},r_{c})$. Half-spaces made by the rays are represented as $\mathcal{H}_{o2}^{ij}$ and $\mathcal{H}_{o2}^{ji}$. To satisfy the conditions that $V_{oi}$ and $V_{oj}$ are on $\overline{V_{i}V_{Q}}$ and $\overline{V_{j}V_{Q}}$, and the rays do not intersect, $\alpha_{o}^{ij}$ should be in the following range.
\begin{equation}
    \label{subeq:obtuse_alpha_range}
    \begin{aligned}
        &1\leq \alpha_{o}^{ij} \leq \min\Bigg(\frac{\min(\|\textbf{x}_{cq0}^{i}\|, \|\textbf{x}_{cq0}^{j}\|)}{r_{c}}, \sqrt{\frac{2}{1-(\textbf{n}_{cq0}^{i})^{\top}\textbf{n}_{cq0}^{j}}}\Bigg)\\
        &\text{where}\  \textbf{x}_{cq0}^{i}=\textbf{x}_{c0}^{i}-\textbf{x}_{q0},\ \textbf{x}_{cq0}^{j}=\textbf{x}_{c0}^{j}-\textbf{x}_{q0},\\
        & \quad \quad \ \
        \textbf{n}_{cq0}^{i}=\textbf{x}_{cq0}^{i}/\|\textbf{x}_{cq0}^{i}\|,\ \textbf{n}_{cq0}^{j}=\textbf{x}_{cq0}^{j}/\|\textbf{x}_{cq0}^{j}\|
    \end{aligned}
\end{equation}
With the $\alpha_{o}^{ij}$ satisfying (\ref{subeq:obtuse_alpha_range}), for $\forall\textbf{x}^{i} \in \mathcal{H}_{o1}^{ij}\cap\mathcal{H}_{o2}^{ij}$ and $\forall\textbf{x}^{j} \in \mathcal{H}_{o1}^{ji}\cap\mathcal{H}_{o2}^{ji}$, the following inequalities are satisfied.
\begin{equation}
    \label{eq:mvc_line_of_sight_ok}
    \begin{aligned}
    &\underset{\epsilon\in[0,1]}{\min} \|\epsilon\textbf{x}^{i}+(1-\epsilon)\textbf{x}_{q0}-\textbf{x}^{j}\|> r_{c},\\  
    &\underset{\epsilon\in[0,1]}{\min} \|\epsilon\textbf{x}^{j}+(1-\epsilon)\textbf{x}_{q0}-\textbf{x}^{i}\|> r_{c} 
    \end{aligned}
\end{equation}
The above inequalities mean that the distances between neighboring agents and \textit{Line-of-Sight} connecting the tracker and the target are always greater than the size of the trackers $r_{c}$; therefore, the inter-occlusion constraints are satisfied. To make the inter-occlusion-free regions move with the target, we maintain the shapes of $\mathcal{H}_{o1}^{ij}\cap\mathcal{H}_{o2}^{ij}$ and $\mathcal{H}_{o1}^{ji}\cap\mathcal{H}_{o2}^{ji}$ and translate them by $\textbf{x}_{q}(t)-\textbf{x}_{q0}$, equivalent to the amount of target's movement. The moving half-spaces are represented as $\mathcal{H}_{o1}^{ij}(t)$, $\mathcal{H}_{o2}^{ij}(t)$, $\mathcal{H}_{o1}^{ji}(t)$, and $\mathcal{H}_{o2}^{ji}(t)$. Their mathematical expressions are simplified as follows, and $\mathcal{H}_{o1}^{ji}(t)$ and $\mathcal{H}_{o2}^{ji}(t)$ can be constructed by exchanging the indices $i$ and $j$ in (\ref{eq:mvc_obtuse}).
\begin{subequations}
\label{eq:mvc_obtuse}
    \begin{align}
    \label{subeq:mvc_obtuse_first_i}
    \mathcal{H}_{o1}^{ij}(t) = & \{\textbf{x}(t)\in\mathbb{R}^{2}|(\textbf{n}_{cq0}^{i})^{\top}(\textbf{x}(t)-\textbf{x}_{q}(t))-\alpha_{o}^{ij} r_{c} \geq 0 \},\\
    \label{subeq:mvc_obtuse_second_i} \nonumber
    \mathcal{H}_{o2}^{ij}(t)= & \bigg \{ \textbf{x}(t)\in \mathbb{R}^{2}| \begin{bmatrix}
       z^{ij}\sin({\theta}_{cq}^{j}+z^{ij}\theta_{o}^{ij})\\
       -z^{ij}\cos({\theta}_{cq}^{j}+z^{ij}\theta_{o}^{ij})
    \end{bmatrix}^{\top} (\textbf{x}(t)- \\ & \ \textbf{x}_{q}(t)-\alpha_{o}^{ij} r_{c}\textbf{n}_{cq0}^{i}) \leq 0 \bigg \}
    \end{align}
\end{subequations}
where $z^{ij}$, $\theta_{cq}^{i}$, $\theta_{cq}^{j}$, and $\theta_{o}^{ij}$ are defined as follows.
\begin{equation}
\label{eq:mvc_obtuse_parameters}
    \begin{aligned}
        &z^{ij} = \text{sign}\big(\det\big(
            [(\textbf{x}_{cq0}^{j})^{\top};(\textbf{x}_{cq0}^{i})^{\top}]
        \big)\big),\\
        &\begin{bmatrix}
            \cos(\theta_{cq}^{i})\\ \sin(\theta_{cq}^{i})
        \end{bmatrix} = \textbf{n}_{cq0}^{i},\ \begin{bmatrix}
            \cos(\theta_{cq}^{j})\\ \sin(\theta_{cq}^{j})
        \end{bmatrix} = \textbf{n}_{cq0}^{j},\\
        &\sin(\theta_{o}^{ij}) =(\alpha_{o}^{ij})^{-1},\ \cos(\theta_{o}^{ij})=\sqrt{1-\sin^{2}(\theta_{o}^{ij})}
    \end{aligned}
\end{equation}
\subsubsection{Acute case}
\label{subsubsec:mvc_acute}
Fig. \ref{fig:constraint_mvc_acute} illustrates an acute case.
First, we draw the lines that are parallel to  $\overline{V_{i}V_{Q}}$ and $\overline{V_{j}V_{Q}}$ and apart by $\alpha_{a}^{ij}r_{c}$ ($\alpha_{a}^{ij}\geq1$) from them. The half-spaces made by the lines are represented as $\mathcal{H}_{a1}^{ji}$ and $\mathcal{H}_{a1}^{ij}$ (red in Fig. \ref{fig:constraint_mvc_acute}), and the intersections between the segments and lines are denoted as $V_{aj}$ and $V_{ai}$, respectively. Then, we draw the rays (blue in Fig. \ref{fig:constraint_mvc_acute}) starting from $V_{ai}$ and $V_{aj}$, which are parallel to the lines that pass through $V_{Q}$ and are tangential to the balls $\mathcal{B}(V_{aj},r_{c})$ and $\mathcal{B}(V_{ai},r_{c})$. Half-spaces divided by the rays are represented as $\mathcal{H}_{o2}^{ij}$ and $\mathcal{H}_{o2}^{ji}$. To satisfy conditions that $V_{ai}$ and $V_{aj}$ are on $\overline{V_{i}V_{Q}}$ and $\overline{V_{j}V_{Q}}$, and the rays do not intersect, $\alpha_{a}^{ij}$ should satisfy the following range.
\begin{equation}
    \label{eq:acute_alpha_range}
\begin{aligned}
&1\leq \alpha_{a}^{ij} \leq \min\bigg(\frac{\big\|\det([
            (\textbf{x}_{cq0}^{i})^{\top};(\textbf{x}_{cq0}^{j})^{\top}]
         )\big\|}{r_{c}\max(\|\textbf{x}_{cq0}^{i}\|,\|\textbf{x}_{cq0}^{j}\|)},\\
& \quad \quad \quad  \quad \quad \quad \quad \quad \quad\sqrt{2(1+(\textbf{n}_{cq0}^{i})^{\top}\textbf{n}_{cq0}^{j})} \bigg)
\end{aligned}
\end{equation}
Similarly to the \textit{Obtuse case}, $\forall\textbf{x}^{i} \in \mathcal{H}_{a1}^{ij}\cap\mathcal{H}_{a2}^{ij}$ and $\forall\textbf{x}^{j} \in \mathcal{H}_{a1}^{ji}\cap\mathcal{H}_{a2}^{ji}$ satisfy (\ref{eq:mvc_line_of_sight_ok}). The moving version of these half-spaces are denoted as $\mathcal{H}_{a1}^{ij}(t)$, $\mathcal{H}_{a2}^{ij}(t)$, $\mathcal{H}_{a1}^{ji}(t)$, and $\mathcal{H}_{a2}^{ji}(t)$ and mathematically formulated as follows.
\begin{subequations}\label{eq:mvc_acute}
    \begin{align}
        \label{subeq:acute_first_i}
        \nonumber \mathcal{H}^{ij}_{a1}(t)= &\big\{ \textbf{x}(t)\in \mathbb{R}^{2}| z^{ij}\det\big([(\textbf{x}(t)-\textbf{x}_{q}(t))^{\top};
        (\textbf{x}_{cq0}^{j})^{\top}]\big)+\\ &\quad \quad \quad \alpha_{a}^{ij}r_{c}\|\textbf{x}_{cq0}^{j}\| \leq 0 \big\},\\
        \label{subeq:acute_second_i} \nonumber 
        \mathcal{H}^{ij}_{a2}(t)= &\bigg\{ \textbf{x}(t) \in \mathbb{R}^{2}| \begin{bmatrix}
    z^{ij}\sin(\theta_{cq}^{j}+z^{ij}\theta_{a}^{ij})\\
            -z^{ij}\cos(\theta_{cq}^{j}+z^{ij}\theta_{a}^{ij})
        \end{bmatrix}^{\top}\bigg(\textbf{x}(t)-\\ &\ \textbf{x}_{q}(t)-\frac{\alpha_{a}^{ij}r_{c}\textbf{n}_{cq0}^{i}}{\big\|\det \big([(\textbf{n}_{cq0}^{i})^{\top};(\textbf{n}_{cq0}^{j})^{\top}\big)\big\|} \big)\leq 0 \bigg\}
    \end{align}
\end{subequations}
where $\theta_{a}^{ij}$ are defined as follows.
\begin{equation}
\label{eq:mvc_acute_parameters}
    \begin{aligned}
        &\sin(\theta_{a}^{ij}) = \frac{1}{\alpha_{a}^{ij}}\big\|\det\big(
            [(\textbf{n}_{cq0}^{i})^{\top};(\textbf{n}_{cq0}^{j})^{\top}]
        \big)\big\|,\\
        &\cos(\theta_{a}^{ij}) = \sqrt{1-\sin^{2}(\theta_{a}^{ij})}
    \end{aligned}
\end{equation}
\begin{lemma}
\label{th:mbvc_and_mvc}
The intersection of DBVC and DIVC is non-empty.
\end{lemma}
\begin{proof} By moving the $i$-th agent by the same relative displacement as the target's movement: $\textbf{x}_{c}^{i}(t)=\textbf{x}_{c0}^{i}+(\textbf{x}_{q}(t)-\textbf{x}_{q0})$, both constraints $\textbf{x}_{c}^{i}(t)\in \mathcal{H}_{s}^{ij}(t)$ and $\textbf{x}_{c}^{i}(t)\in \mathcal{H}_{\mu1}^{ij}(t)\cap\mathcal{H}_{\mu2}^{ij}(t), \mu=o\ \text{or}\ a$, are satisfied.  
\end{proof}
\textbf{Lemma} \ref{th:mbvc_and_mvc} indicates that the DBVC and DIVC offer feasible constraints to handle inter-agent collisions and occlusions.
\section{Tracking Trajectory Planning}
\label{sec:tracking_trajectory_planning}
In this section, we formulate the tracking trajectory generation using Bernstein polynomial motion primitives. We begin by sampling a set of motion primitives and filtering out those that do not satisfy the given constraints. Next, we select the optimal trajectory. To enhance computational efficiency during both the feasibility check and trajectory selection processes, we fully exploit the convex hull property and the integral property of the Bernstein polynomial, respectively.
\subsection{Primitive Sampling}
\label{subsec:primitive_sampling}
Based on the $i$-th agent's current position $\textbf{x}_{c0}^{i}$ and velocity $\dot{\textbf{x}}_{c0}^{i}$, we formulate the following optimal control problem to sample primitives, which can be solved in the closed form. 
\begin{equation}
    \label{eq:nearly_constant_velocity_model}
    \begin{aligned}
    &\underset{\textbf{u}_{c}^{i}(t)}{\text{min}} && \frac{1}{T}\int_{0}^{T}\|\textbf{u}_{c}^{i}(t)\|^{2}dt \\
    &\text{s.t.} &&  \dot{\textbf{z}}_{c}^{i}(t)
     = F_{c}\
    \textbf{z}_{c}^{i}(t) + G_{c}\textbf{u}_{c}^{i}(t),\\
     & \ && \textbf{z}_{c}^{i}(t)=\begin{bmatrix}
         \textbf{x}_{c}^{i}(t) \\ \dot{\textbf{x}}_{c}^{i}(t)
     \end{bmatrix}, F_{c} =
\begin{bmatrix}
    0_{2} & I_{2}\\
    0_{2} & 0_{2}
\end{bmatrix}
, \ G_{c} = 
\begin{bmatrix}
    0_{2} \\
    I_{2}
\end{bmatrix},\\
& \ && \textbf{x}_{c}^{i}(0)= \textbf{x}_{c0}^{i}, \ \dot{\textbf{x}}_{c}^{i}(0) = \dot{\textbf{x}}_{c0}^{i}, \ \textbf{x}_{c}^{i}(T)=\textbf{x}_{cf}^{i}
    \end{aligned}
\end{equation}
$0_{2}$ is a $2\times2$ zero matrix, and $I_{2}$ is a $2 \times 2$ identity matrix. The $\textbf{x}_{cf}^{i}$ are terminal points, sampled around the target's terminal points $\textbf{x}_{q}(T)$:

\begin{equation}
    \label{eq:terminal_point_sampling}
    \begin{aligned}
    &\textbf{x}_{cf}^{i} = \textbf{x}_{q}(T)+[r_{cs}^{i}\cos{\psi_{cs}^{i}},r_{cs}^{i}\sin{\psi_{cs}^{i}} ]^{\top},\\
    &r_{cs}^{i} \sim U[\underbar{$r$}_{cs}^{i},\bar{r}_{cs}^{i}],\ \psi_{cs}^{i} \sim U [\underbar{$\psi$}_{cs}^{i},\bar{\psi}_{cs}^{i}].    
    \end{aligned}
\end{equation}
$U$ represents the uniform distribution, and $(\underbar{$r$}_{cs}^{i},\bar{r}_{cs}^{i})$ and $(\underbar{$\psi$}_{cs}^{i},\bar{\psi}_{cs}^{i})$ are pairs of the lower and upper bound of distribution of radius and azimuth, respectively.
\subsection{Feasibility Check}
\label{subsec:feasibility_check}
\subsubsection{Collision Check}
\label{subsubsec:collision_check}
To make the $i$-th agent safe, we check whether trajectories satisfy (\ref{subeq:collision_between_static}): being confined in safe and visible corridors, and (\ref{subeq:collision_between_dynamic})-(\ref{subeq:collision_between_agent}): avoiding collision with dynamic obstacles, target, and the other agents. 
\begin{subequations}
    \label{eq:collision_check}
\begin{align}
    \label{subeq:collision_between_static}
    & \Xi(\textbf{C}_{m}^{i})\oplus \mathcal{B}(\textbf{0},r_{c})\subset \mathcal{S}_{m}^{i},\ \forall m=1,\ldots,M_{i}, \\
    \label{subeq:collision_between_dynamic}
    & \|\textbf{x}_{c}^{i}(t)-\textbf{x}_{o}^{k}(t)\|^{2}-(r_{c}+r_{o})^{2}\geq 0,\ \forall k\in \mathcal{I}_{o}\\
    \label{subeq:collision_between_target}
    & \|\textbf{x}_{c}^{i}(t)-\textbf{x}_{q}(t)\|^{2}- (r_{c}+r_{q})^{2}\geq0,\\
    \label{subeq:collision_between_agent}
    & \textbf{x}_{c}^{i}(t) \in \bigcap_{j\in\mathcal{I}_{c}\setminus i} \mathcal{H}_{s}^{ij}(t),
\end{align}
\end{subequations}
$\mathcal{S}_{m}^{i}$ is the $m$-th visible and safe corridors generated by \cite{bpmp-tracker}, $\textbf{C}_{m}^{i}$ is the $m$-th Bernstein coefficients, which is split from $\textbf{C}^{i}$. $\Xi$ is a set of points in Bernstein coefficients.  
For (\ref{subeq:collision_between_static}), we utilize convex hull property, and since the left-hand-side of (\ref{subeq:collision_between_dynamic}) and (\ref{subeq:collision_between_target}) can be expressed as Bernstein polynomials, we select primitives with nonnegative coefficients for all bases.
Similarly to (\ref{subeq:collision_between_static}), we verify whether the control points of (\ref{subeq:collision_between_agent}) belong to the intersections of the affine spaces by using the convex hull property to determine if trajectories are located within DBVC where inter-collision does not occur. The primitives that pass the tests in (\ref{eq:collision_check}) satisfy (\ref{eq:collision_avoidance}).
\subsubsection{Visibility Check}
\label{subsubsec:visibility_check}
\begin{figure}
    \centering
    \includegraphics[width=0.4\linewidth]{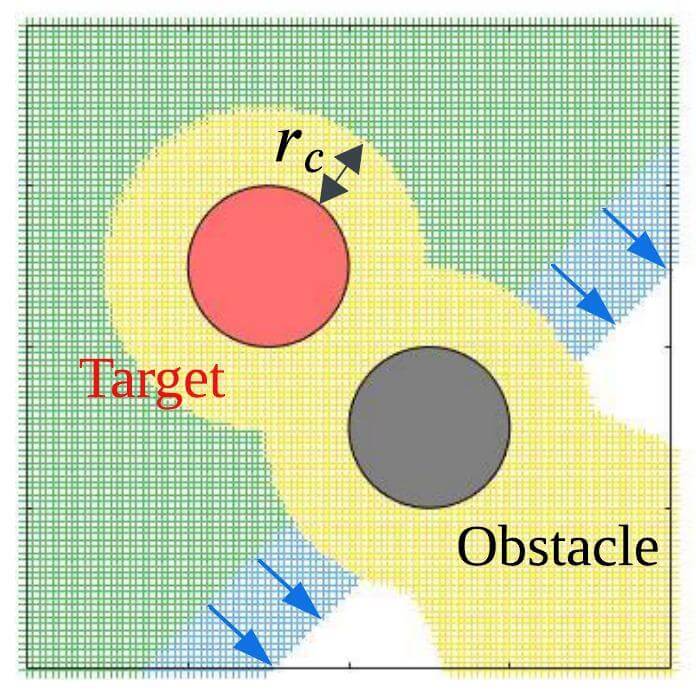}
    \caption{Comparison of the feasibility checks. Yellow: an area where the agent either cannot see the center of a target or collides with the target or an obstacle. Green: an area where \cite{bpmp-tracker} considers the collision- and occlusion-free area. Blue: an expanded feasible area by the proposed method.}
    \label{fig:conservativeness_comparison}
    \vspace{-4mm}
\end{figure}

We check whether the primitives of the $i$-th agent satisfy (\ref{subeq:vis_static}), (\ref{subeq:vis_dynamic}), and (\ref{subeq:vis_inter}) to avoid occlusion by static obstacles, dynamic obstacles, and the other agents, respectively, and consequently satisfy (\ref{eq:occlusion_avoidance}).
\begin{subequations}
    \label{eq:occlusion_check}
    \begin{align}
        \label{subeq:vis_static}
        & \Xi(\textbf{C}_{m}^{i}) \subset \mathcal{S}_{m}^{i},\ \forall 
 k=1,\ldots,M_{i},\\
        \label{subeq:vis_dynamic}
        & \|\epsilon\textbf{x}_{c}^{i}(t)+(1-\epsilon)\textbf{x}_{q}(t)-\textbf{x}_{o}^{k}(t)\|^{2}-r_{o}^{2}\geq 0, \\ \nonumber
        & \forall k\in \mathcal{I}_{o},\ \forall \epsilon \in [0,1],\\
        \label{subeq:vis_inter}
        &\textbf{x}_{c}^{i}(t)\in \bigcap_{j\in\mathcal{I}_{c}\setminus i} \big(\mathcal{H}_{\mu1}^{ij}(t) \cap \mathcal{H}_{\mu2}^{ij}(t) \big),\ \mu = o\ \text{or}\ a
\end{align}
\end{subequations}
(\ref{subeq:vis_static}) is a necessary condition for (\ref{subeq:collision_between_static}). The convex hull property is used to check (\ref{subeq:vis_inter}), the DIVC constraints.
(\ref{subeq:vis_dynamic}) means that \textit{Line-of-Sight} does not collide with dynamic obstacles, and the left-hand side of (\ref{subeq:vis_dynamic}) can be reformulated as follows.
\begin{equation}
    \begin{aligned}
    &\text{L.H.S of (\ref{subeq:vis_dynamic})}=
    \epsilon^{2}\sigma_{1}(t)+2\epsilon(1-\epsilon)\sigma_{2}(t)+(1-\epsilon)^{2}\sigma_{3}(t)\\
    & \quad \quad \quad \quad \quad \quad \geq\epsilon^{2}\sigma_{1}'(t)+2\epsilon(1-\epsilon)\sigma_{2}'(t)+(1-\epsilon)^{2}\sigma_{3}'(t),\\
    &\text{where}\ \sigma_{1}(t)= \|\textbf{x}_{c}^{i}(t)-\textbf{x}_{o}^{k}(t)\|^{2}-r_{o}^{2},\\
    &\quad \quad \ \ \sigma_{2}(t)=(\textbf{x}_{c}(t)-\textbf{x}_{o}^{k}(t))^{\top}(\textbf{x}_{q}(t)-\textbf{x}_{o}^{k}(t))-r_{o}^{2},\\
    &\quad \quad \ \ \sigma_{3}(t)=\|\textbf{x}_{q}(t)-\textbf{x}_{o}^{k}(t)\|^{2}-r_{o}^{2},\\
    &\quad \quad \ \ \sigma_{1}'(t)= \sigma_{1}(t)-(r_{o}+r_{c})^{2}+r_{o}^{2},\\
    &\quad \quad \ \ \sigma_{2}'(t)=\sigma_{2}(t)+(r_{o}+\min(r_{q},r_{c}))^{2}-r_{o}^{2},\\
    &\quad \quad \ \ \sigma_{3}'(t)=\sigma_{3}(t)-(r_{o}+r_{q})^{2}+r_{o}^{2}
    \end{aligned}
\end{equation}
Since $\epsilon^{2}$, $2\epsilon(1-\epsilon)$, and $(1-\epsilon)^{2}$ are nonnegative for $\forall \epsilon\in[0,1]$, (\ref{subeq:vis_dynamic}) holds if the terms multiplied by them are also nonnegative. This paper examines whether $\sigma_{1}'(t)$, $\sigma_{2}'(t)$, $\sigma_{3}'(t)\geq0$ hold, whereas our previous work \cite{bpmp-tracker} checks whether $\sigma_{1}(t)$, $\sigma_{2}(t)$, $\sigma_{3}(t)\geq0$ hold.
By tightening the conditions for $\sigma_{1}(t)$ and $\sigma_{3}(t)$, the proposed work relaxes condition for $\sigma_{2}(t)$.
Fig. \ref{fig:conservativeness_comparison} shows that the proposed check method is less conservative compared to \cite{bpmp-tracker}. By employing the less conservative feasibility check, our planner can find more feasible motions.
\subsubsection{Distance Check}
\label{subsubsec:distance_check}
To keep suitable distance from the targets, the following conditions are established.
\begin{equation}
    \label{eq:distance_target_divded}
    \|\textbf{x}_{c}^{i}(t)-\textbf{x}_{q}(t)\|^{2}-d_{\min}^{2}\geq 0,\ d_{\max}^{2}-\|\textbf{x}_{c}^{i}(t)-\textbf{x}_{q}(t)\|^{2}\geq 0
\end{equation}
We set $d_{\min}\geq r_{q}+r_{c}$ to avoid collision with the target, and the primitives that pass the tests in (\ref{eq:distance_target_divded}) satisfy (\ref{eq:distance_target}). 
\begin{table*}[t!]
    \vspace{2mm}
    \centering
    \caption{Reported Validation Performance}
    \label{tab:reported_performance}
    \begin{tabular}{ccc|ccc|cc}
    \toprule
      \multirow{2}{*}{Scenarios}&\multirow{2}{*}{Environments} & \multirow{2}{*}{Planner}&\multicolumn{3}{c}{\textbf{Safety Metrics [m]}} &\multicolumn{2}{c}{\textbf{Visibility Metrics [m]}}\\
        & & & Obstacle (\ref{subeq:distance_metric_obstacle}) & Agent (\ref{subeq:distance_metric_inter}) & Target (\ref{subeq:distance_metric_target}) & Obstacle (\ref{subeq:vis_metric_obstacle}) & Agent (\ref{subeq:vis_metric_inter})\\
     \hline
      \multirow{2}{*}{Sc1}&\multirow{2}{*}{unstructured}& proposed &0.609/0.132 &0.561/0.727 &0.215/0.359 &0.202/0.609 &0.290/0.727\\
     & &baseline \cite{fei-gao-wild} &0.098/0.567 &0.225/0.639 &\textbf{0.000}/0.399 &0.173/0.567 &\textbf{0.000}/0.639\\
     Sc2 &dynamic& proposed &0.061/0.327 &0.480/0.666 &0.195/0.319 &0.136/0.327 &0.270/0.666\\
     Sc3 &unstructured& proposed &0.013/0.480 &0.554/0.683 &0.148/0.346 &0.089/0.480 &0.223/0.683\\
     Sc4 &dynamic& proposed &0.065/1.201 &0.421/0.711 &0.240/0.353 &0.140/1.201 &0.315/0.711\\
     Sc5 &unstructured& proposed &0.413/1.014 &0.467/0.848 &0.275/0.654 &0.401/1.014 &0.319/0.848\\
     Sc6 &dynamic& proposed &0.135/0.397 &0.378/0.431 &0.096/0.217 &0.210/0.397 &0.171/0.431\\
    \hline
    \end{tabular}
     \vspace{1mm}
     \begin{minipage}{\textwidth}
     \textbf{0} indicates a collision or occlusion. The values for the above metrics indicate the minimum/mean performance.
     \end{minipage}
     \vspace{-6mm}
\end{table*}
\subsubsection{Dynamical Limit Check}
\label{subsubsec:dynamical_limit_check}
To ensure the drone does not exceed dynamic limits (\ref{eq:dynamical_limits}), we verify whether the primitives satisfy the following inequalities. (\ref{subeq:dynamical_limit_rotation1}) is calculated under the assumption that the agents directly head toward the target.
\begin{subequations}
    \label{eq:dynamical_limit_summary}
    \begin{align}
        \label{subeq:vel_max}
        &v_{\max}^{2}-\|\dot{\textbf{x}}_{c}^{i}(t)\|^{2} \geq 0,\ a_{\max}^{2}-\|\ddot{\textbf{x}}_{c}^{i}(t)\|^{2} \geq 0,\\
        \label{subeq:dynamical_limit_rotation1}
        &\dot{\psi}_{c}(t) =
        \frac{
        \det \big(
            [(\textbf{x}_{q}(t)-\textbf{x}_{c}^{i}(t))^\top;
            (\dot{\textbf{x}}_{q}(t)-\dot{\textbf{x}}_{c}^{i}(t))^\top] \big)}
        {\|\textbf{x}_{q}(t)-\textbf{x}_{c}^{i}(t)\|^{2}},\\
        \label{subeq:dynamical_limit_rotation2}
        & -\dot{\psi}_{\max} \leq \dot{\psi}_{c}(t) \leq \dot{\psi}_{\max}
    \end{align}
\end{subequations}
\subsection{Best Trajectory Selection}
\label{subsec:best_trajectory_selection}
Among the primitives that pass (\ref{eq:collision_check})-(\ref{eq:dynamical_limit_summary}), we select the best tracking trajectory. We evaluate the following cost, which consists of a penalty term for jerkiness of trajectory and a cost term to maintain appropriate distance from the target.
\begin{subequations}
    \label{eq:chasing_cost}
    \begin{align}
        \min \ &J_{1}+J_{2}, \\
        &J_{1} = w_{j}\int_{0}^{T}\|\dddot{\textbf{x}}_{c}^{i}(t)\|_{2}^{2}dt,\\
        &J_{2} = \sum_{i=1}^{N_{q}} \int_{0}^{T}(\|\textbf{x}_c^{i}(t)-\textbf{x}_{q}(t)\|_{2}^{2}-d_{des,i}^{2})^{2}dt
    \end{align}
\end{subequations}
$w_{j}$ is a weight factor, and $d_{des,i}$ is the desired distance between the target and the $i$-th agents, which is set to $\frac{1}{2}(\underbar{$r$}_{cs}^{i}+\bar{r}_{cs}^{i})$ in the validation.
\section{Validations}
\label{sec:validations}
In this section, the proposed method is validated through
various target-tracking settings. We measure the distance
between the drone and environments, defined as (\ref{subeq:distance_metric_obstacle}), the distance among agents (\ref{subeq:distance_metric_inter}), and the distance between the target and drone, defined as (\ref{subeq:distance_metric_target}), to evaluate drone safety.
\begin{subequations}
    \label{eq:metrics_safety}
    \begin{align}
    \label{subeq:distance_metric_obstacle}
        &\chi_{1}(t) = \min_{i\in\mathcal{I}_{c}}\min_{\substack{\textbf{x}(t)\in\mathcal{C}_{i}(t)\\\textbf{y}(t)\in\mathcal{O}(t)}}\|\textbf{x}(t)-\textbf{y}(t)\|, \\
        \label{subeq:distance_metric_inter}
        & \chi_{2}(t) = \min_{i\in\mathcal{I}_{c}}\min_{\substack{\textbf{x}(t)\in\mathcal{C}_{i}(t)\\\textbf{y}(t)\in\mathcal{C}^{i}(t)}}\|\textbf{x}(t)-\textbf{y}(t)\|,\\
        \label{subeq:distance_metric_target}
        & \chi_{3}(t) = \min_{i\in\mathcal{I}_{c}}\min_{\substack{\textbf{x}(t)\in\mathcal{C}_{i}(t)\\\textbf{y}(t)\in\mathcal{T}(t)}}\|\textbf{x}(t)-\textbf{y}(t)\|
    \end{align}
\end{subequations}
Also, we measure the distance between the \textit{Lines-of-Sight} and environments (\ref{subeq:vis_metric_obstacle}) and the distance between the \textit{Lines-of-Sight} and the other agents (\ref{subeq:vis_metric_inter}) to assess the target visibility.
\begin{subequations}
    \label{eq:metrics_visibility}
    \begin{align}
        \label{subeq:vis_metric_obstacle}
        &\phi_{1}(t)= \min_{i\in \mathcal{I}_{c}}\min_{\substack{\textbf{x}(t)\in \mathcal{L}(\textbf{x}_{c}^{i}(t),\textbf{x}_{q}(t))\\ \textbf{y}(t)\in \mathcal{O}(t)}}\|\textbf{x}(t)-\textbf{y}(t)\|,\\
        \label{subeq:vis_metric_inter}
        &\phi_{2}(t)= \min_{i\in \mathcal{I}_{c}}\min_{\substack{\textbf{x}(t)\in \mathcal{L}(\textbf{x}_{c}^{i}(t),\textbf{x}_{q}(t))\\ \textbf{y}(t)\in \mathcal{C}^{i}(t)}}\|\textbf{x}(t)-\textbf{y}(t)\|
    \end{align}
\end{subequations}

Using the above performance metrics, we validate the operability of the proposed planner under two environmental conditions: 1) unstructured static obstacles, and 2) dynamic obstacles. Through simulations and hardware experiments, we show successful target tracking. Table \ref{tab:reported_performance} shows the reported performance, and the details of the tests are explained in the following subsections. In addition, we validate the effectiveness of the proposed method for multi-agent tracking through a comparative analysis and show applicability to 3D scenarios. 

For the target trajectory prediction, we employ a method in \cite{bpmp-tracker}, applicable to both test conditions: environments with unstructured-but-static obstacles and dynamic-but-structured obstacles.
The radii of the targets, trackers, and dynamic obstacles are set to 7.5 cm, matching the size of Crazyflie quadrotors. Also, since the test environments are either narrow or crowded, we set the sampling range of the tracking distance $(\underbar{$r$}_{cs},\bar{r}_{cs})$ to (0.3, 0.6) [m].

In the tests, we use computers with an Intel i7 12th-gen CPU and 16GB RAM. The number of sampled primitives is set to 1000, and four threads are used for parallel computation. The reported computation time in all tracking scenarios is less than 10 milliseconds.
\subsection{Simulations}
\label{subsec:simulations}
\begin{figure}[t!]
\centering
\includegraphics[width = 0.95\linewidth]{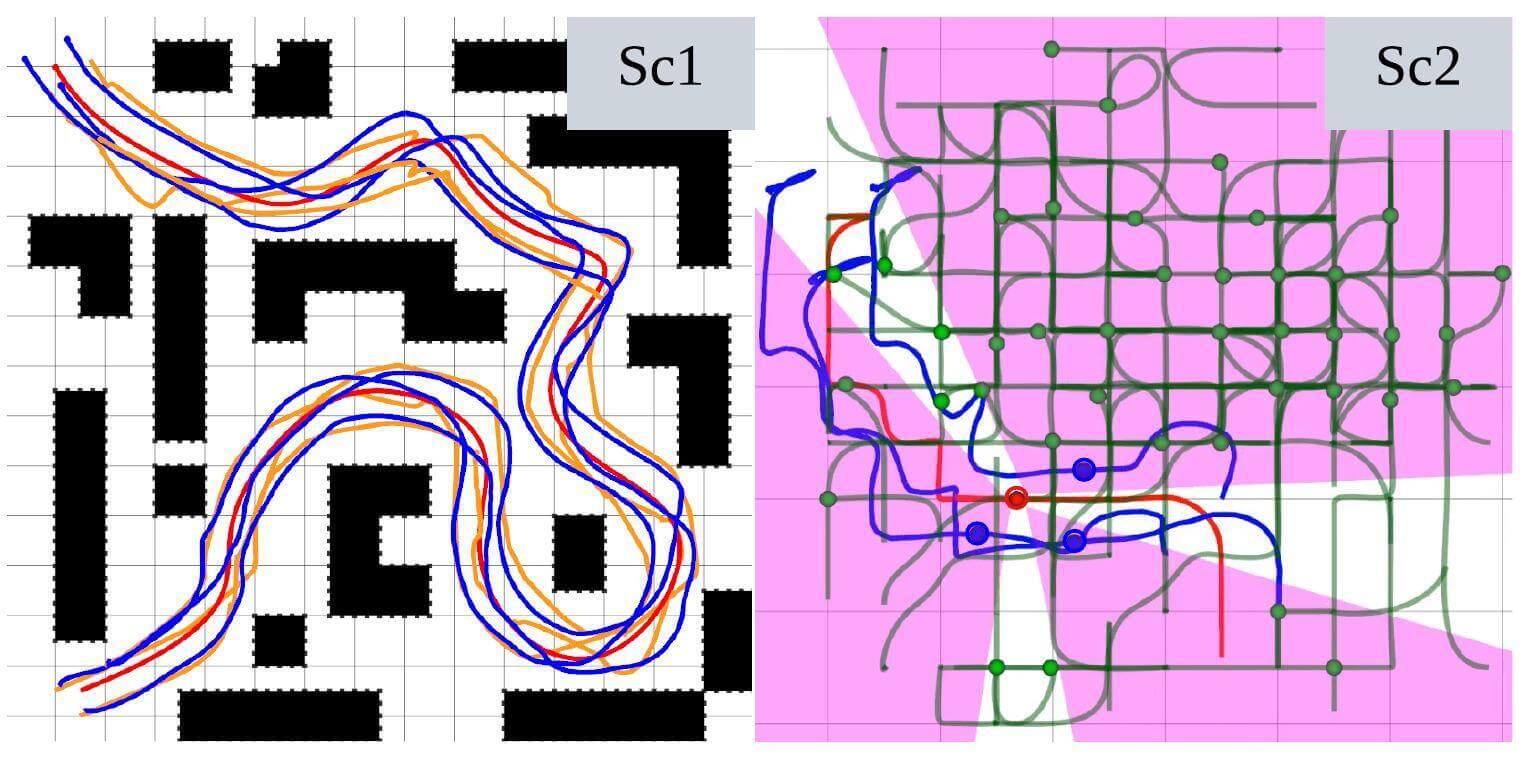}
\caption{\textbf{Left:} Three trackers (blue: ours, orange: \cite{fei-gao-wild}) follow the red target among black static obstacles. \textbf{Right:} Three blue trackers follow the red target among forty green dynamic obstacles. Magenta regions represent DIVCs.}
\label{fig:dmvc_simulation}
\vspace{-4mm}
\end{figure}

\textit{\textbf{Scenario 1 (Unstructured Environment)}}:
We compare the proposed planner with the state-of-the-art planner \cite{fei-gao-wild} in an environment with various shapes of static obstacles, as shown in Fig. \ref{fig:dmvc_simulation}. 
As Table \ref{tab:reported_performance} shows, the trackers controlled by the baseline \cite{fei-gao-wild} experience inter-agent occlusions and collisions with the target several times, despite finely tuned parameters to adapt to the narrow tracking environment. In contrast, our planner successfully tracks the target without collision and occlusion by thoroughly checking safety and visibility constraints, including the tracking distance $(d_{\min},d_{\max})$ and DIVC constraints.

\textit{\textbf{Scenario 2 (Dynamic Environment)}}: We test our approach in a dense moving-obstacle environment. The target moves in a 6 $\times$ 6 $\text{m}^{2}$ space with 40 obstacles. The target and obstacles move around with the maximum speed $1.0\ \text{m/s}$. The right side of Fig. \ref{fig:dmvc_simulation} shows a flight history.
\subsection{Hardware Experiments}
\label{subsec:hardware_experiments}
For hardware demonstration, we use Crazyflie 2.1 quadrotors. One serves as the target, three serve as the trackers, and the remainder act as dynamic obstacles. The target and dynamic obstacles are both controlled by a single Intel NUC, and they follow a pre-calculated path generated by \cite{dlsc}. On the other hand, to implement distributed calculation, each tracker is connected to a separate mini PC.

\begin{figure}[t!]
\centering
\vspace{1mm}
\includegraphics[width = 1.0\linewidth]{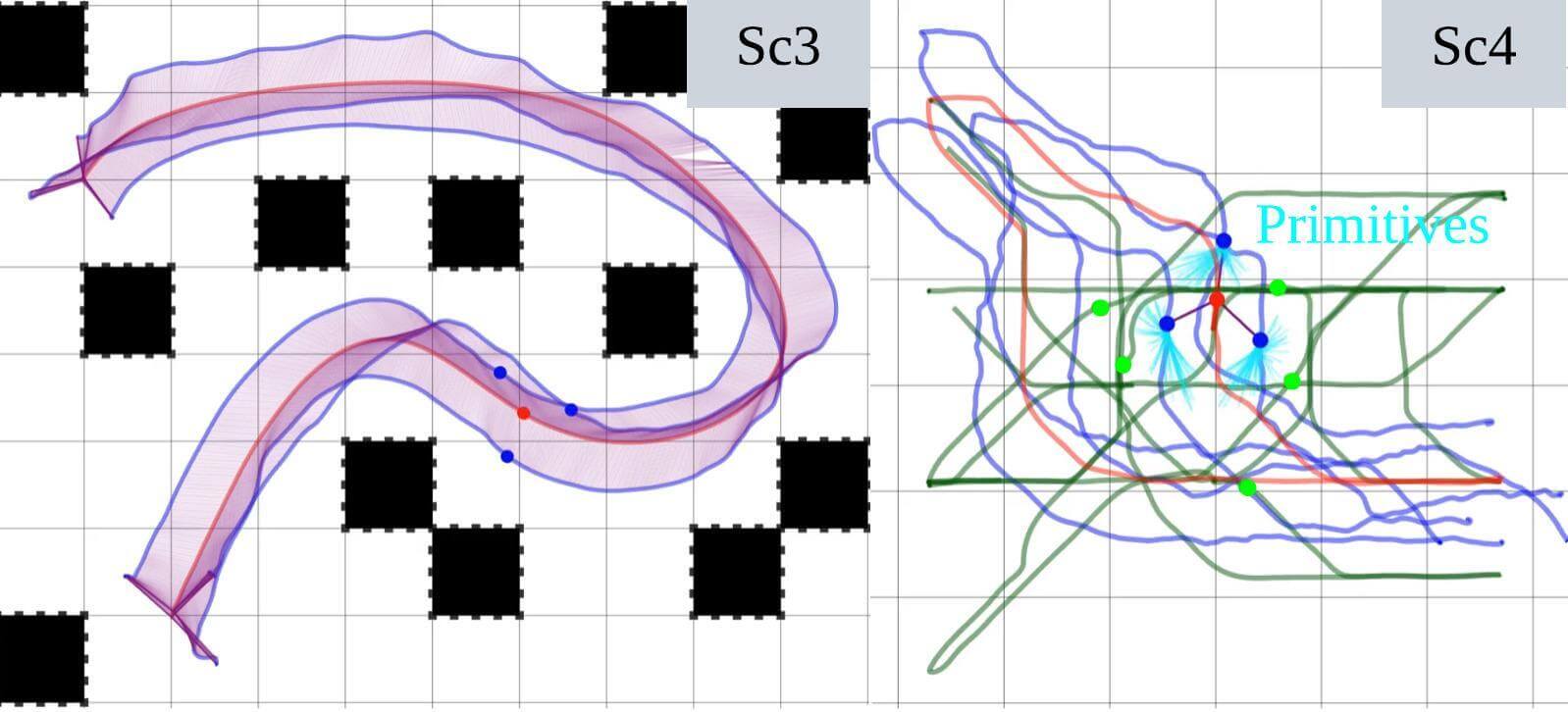}
\caption{Target tracking experiments. The total flight paths of Crazyflies serving as trackers (blue), target (red), and dynamic obstacles (green), along with boxes (black), are plotted in a top-down view. The purple areas are the accumulated histories of the \textit{Lines-of-sight} connecting the target and the trackers.}
\label{fig:dmvc_experiment_rviz}
\vspace{-1mm}
\end{figure}

\textit{\textbf{Scenario 3: (Unstructured Environment)}}
The target moves in a $10\times7$ $\text{m}^{2}$ space with twelve cube-shaped obstacles. The top image in Fig. \ref{fig:thumbnail} is a snapshot of the flight, and the left side of Fig. \ref{fig:dmvc_experiment_rviz} shows the reported results.

\textit{\textbf{Scenario 4: (Dynamic Environment)}}
The target moves in a $7\times7$ $\text{m}^{2}$ space with five moving obstacles. The bottom image in Fig. \ref{fig:thumbnail} is a snapshot of the mission, and the right side of Fig. \ref{fig:dmvc_experiment_rviz} summarizes the flight test.
\subsection{Comparison Analysis}
\label{subsec:compoarison_analysis}
The performance of the proposed approach is investigated through two tests. The tests evaluate the effectiveness of the DBVC, the DIVC, and improved feasibility check methods. In each test, we define success as the absence of occlusion or collision until all objects come to a stop. We conduct 1000 tests for each tracking setup and measure the success rate.

First, we validate the effectiveness of the DBVC and the DIVC by comparing the success rates with cases where these two cells are not applied and where static versions of the cells are used. Specifically, when the cells are not used, each tracker treats the neighbor trackers as dynamic obstacles. In the case of using static cells, we use BVC, $\mathcal{H}_{\mu1}^{ij}$, and $\mathcal{H}_{\mu2}^{ij}, \mu=o$ or $a$, instead of $\mathcal{H}_{s}^{ij}(t)$, $\mathcal{H}_{\mu1}^{ij}(t)$, and $ \mathcal{H}_{\mu2}^{ij}(t)$. 

We test these setups in an obstacle-free space, where the target moves around at the maximum speed 1.0 m/s for an average of 30 seconds. We varied the number of trackers and distance to the targets, and Table \ref{tab:ablation_studies} summarizes the results. As the number of targets increases and the tracking distance decreases, the difficulty of tracking increases. Without using cells, the level of interference due to the movement of neighboring agents increases as the tracking distance shortens, leading to a higher failure rate. In contrast, the spatially separated characteristics of the DBVC and the DIVC implicitly make a balanced formation, resulting in a higher success rate. Also, static cells can conflict with other constraints, particularly the distance between the target and the trackers. However, the dynamic properties of the DBVC and the DIVC maintain consistent tracking distance, which results in a higher success rate.
Moreover, a higher number of trackers narrows the cells, bringing the agents closer to the cell boundaries.
Such conditions make it difficult to satisfy all the constraints. However, even in the short tracking distance, the planner using the DBVC and the DIVC achieves the highest success rate because the dynamic properties result in fewer constraint violations under such conditions.
\begin{table}[t]
    \centering
    \caption{Success Rate in an Empty Space When Using (No/Static/Dynamic) Cells [\%]} 
    \label{tab:ablation_studies}
\begin{tabular}{c|ccc}
\hline
$N_{c}$ & 3  & 4 & 5  \\ 
\hline
\begin{tabular}[c]{@{}c@{}}Short\\{[}0.4, 1.2] [m]\end{tabular}  & 95.6/79.4/\textbf{99.5} & 44.7/44.0/\textbf{99.3} & 1.9/14.6/\textbf{96.3}  \\ 
\hline
\begin{tabular}[c]{@{}c@{}}Medium\\{[}0.8, 1.6] [m]\end{tabular} & 97.9/\textbf{99.5}/\textbf{99.5} & 74.2/96.6/\textbf{99.4} & 16.5/67.6/\textbf{98.2}  \\ 
\hline
\begin{tabular}[c]{@{}c@{}}Long\\{[}1.2, 2.0] [m]\end{tabular}    & 98.1/\textbf{99.6}/\textbf{99.6} & 88.0/98.4/\textbf{99.5} & 48.5/96.6/\textbf{98.9} \\
\hline
\end{tabular}
\\
\footnotesize{ The ranges in the first column mean the sampling distance $[\underbar{$r$}_{cs}$,  $\bar{r}_{cs}$]. \\The dynamic cells denote the DBVC and the DIVC.}
\end{table}
\begin{table}[t!]
    \vspace{2mm}
    \begin{center}
    \caption{Success Rate Comparison in Dynamic Environments [\%]}
    \label{tab:success_rate_comparison}
    \begin{tabular}{cc|ccc}
    \toprule
    \multirow{2}{*}{\#Agent ($N_{c}$)} &\multirow{2}{*}{Planner} &\multicolumn{3}{c}{\# Obstacle ($N_{o}$)}\\
    & & 5& 10& 20\\
    \hline
    \multirow{3}{*}{2} & noncooperative &99.2 &98.7 &95.2\\
    & conservative &99.9 &99.9 &99.2\\
    &\textbf{proposed} &\textbf{99.9} & \textbf{99.9} &\textbf{99.4}\\
    \hline
    \multirow{3}{*}{3} & noncooperative &70.1 &68.4 &58.1\\
    & conservative &96.9 &93.5 &88.3\\
    &\textbf{proposed} &\textbf{97.6} &\textbf{94.6} &\textbf{90.1}\\
    \hline
    \multirow{3}{*}{4} & noncooperative &1.01 &0.93 &0.72\\
    & conservative &89.7 &79.6 &68.5\\
    &\textbf{proposed} &\textbf{91.2} &\textbf{82.1} &\textbf{71.2}\\
    \hline
    \end{tabular}
    \end{center}
     \vspace{2mm}
     \footnotesize{noncooperative: feasibility checks in \cite{bpmp-tracker}, without the DBVC and the DIVC. conservative: feasibility checks in \cite{bpmp-tracker}, with the DBVC and the DIVC.}
     \vspace{-4mm}
\end{table}

Second, we conduct a benchmark test to validate the superiority of our planner in a dynamic obstacle environment. We compare our method with a noncooperative approach and a conservative approach. For the noncooperative approach, we use the BPMP-Tracker \cite{bpmp-tracker}. Since the method in \cite{bpmp-tracker} is designed for single-agent tracking, we adapt the problem setting so that each tracker treats neighboring agents as obstacles. The conservative approach utilizes the DBVC and the DIVC but applies the conservative feasibility check methods from \cite{bpmp-tracker}.

The target and obstacles move around for an average of 40 seconds in a 6 $\times$ 6 $\text{m}^{2}$ space at the maximum speed of 0.5 m/s. We measure the success rate while varying the number of obstacles. 
Our approach generates tracking trajectories in a cooperative manner, whereas the noncooperative setup causes consistent interference among trackers.
This results in a higher success rate for the proposed planner for all tracking conditions, as shown in Table \ref{tab:success_rate_comparison}. Moreover, our approach discovers more feasible motions than the conservative approach in tight conditions, leading to a higher success rate.

As the number of trackers increases, the cells become smaller, making it difficult to find a feasible motion within those reduced areas in the presence of adjacent dynamic obstacles. In future work, we plan to design inter-occlusion- and inter-collision-free cells that not only translate but also change shape over time, to enhance robustness against the interference of dynamic obstacles.
\begin{figure}[t!]
\centering
\vspace{1mm}
\includegraphics[width = 1.0\linewidth]{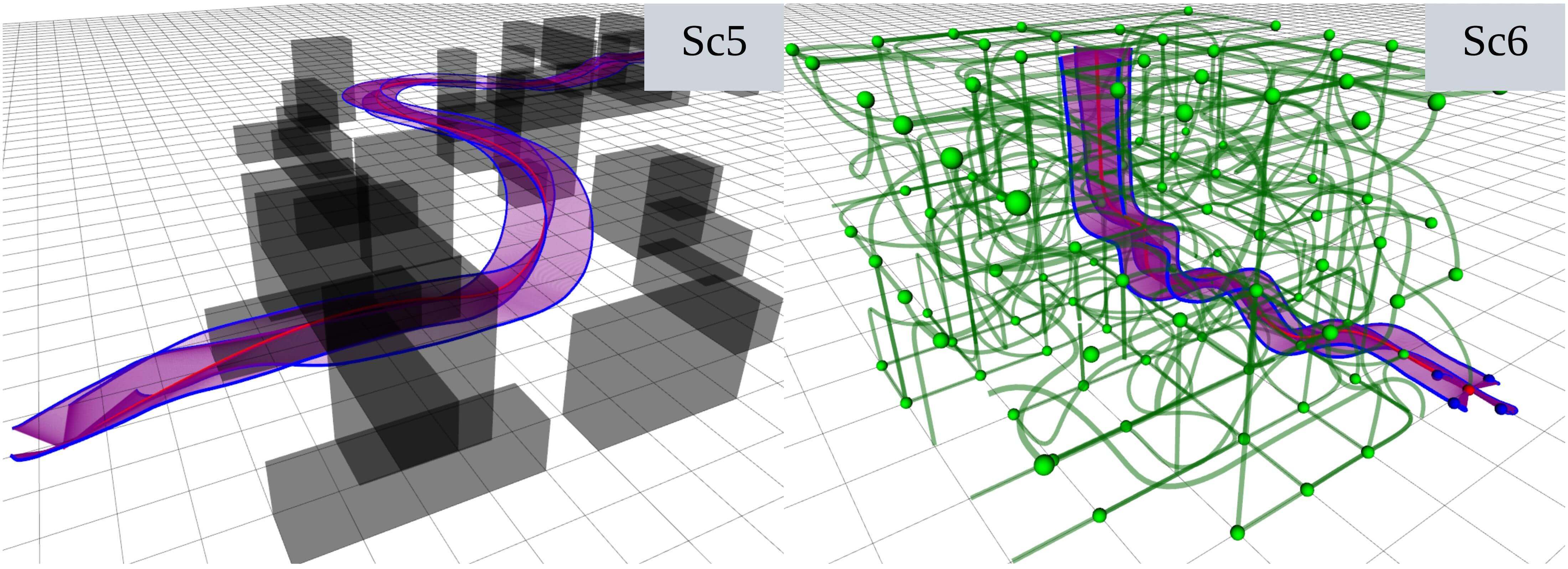}
\caption{Tracking in 3D spaces. Reported paths of trackers (blue), target (red), and dynamic obstacles (green), with static obstacles (black) and \textit{Line-of-sights} (purple) connecting the trackers and the target.}
\label{fig:dmvc_3d_rviz}
\vspace{-1mm}
\end{figure}

\subsection{3D Extensions}
The proposed planner, initially designed for 2D, is extended to 3D environments. Allowing trackers to move along the z-axis enables a wider range of feasible motions and improves tracking performance in challenging scenarios. The following tests show its effectiveness in 3D settings.

\textit{\textbf{Scenario 5: (3D Unstructured Environment)}}
The aerial target navigates though height-varied structures and tunnel-like passages while adjusting its altitude within the range of 1 to 2 meters, with the maximum speed of 2.19 m/s. The left side of Fig. \ref{fig:dmvc_3d_rviz} shows the paths of the target and the four trackers.

\textit{\textbf{Scenario 6: (3D Dynamic Environment)}}
The aerial target flies in a $6\times6\times4$ $\text{m}^{3}$ space with 100 moving airborne obstacles, with the maximum speed of $1.0$ m/s. The right side of Fig. \ref{fig:dmvc_3d_rviz} illustrates the results of tracking using four trackers.

\section{Conclusion}
\label{sec:conclusion}
We presented a distributed multi-agent trajectory generation method for aerial tracking, which prevents both occlusion and collision caused by obstacles. DBVC and DIVC were constructed by dividing the space into multiple cells to avoid inter-agent collision and inter-agent occlusion, respectively.
Since both cells are designed based on the shared current positions of agents and the predicted target's trajectory, it enables distributed planning. We achieved fast computation by combining the DBVC and the DIVC with the Bernstein-polynomial-motion-primitive-based trajectory planner and validated the operability of our planner through tests under various tracking conditions. Lastly, we confirmed that the proposed method achieved a higher success rate in tracking missions in environments with dynamic obstacles, outperforming the state-of-the-art methods.
\bibliographystyle{IEEEtran}
\bibliography{my_bib}
\end{document}